\documentclass{article}

\PassOptionsToPackage{numbers, compress}{natbib}

\usepackage{arxiv}

\usepackage[utf8]{inputenc} %
\usepackage[T1]{fontenc}    %
\usepackage{hyperref}       %
\usepackage{url}            %
\usepackage{booktabs}       %
\usepackage{amsfonts}       %
\usepackage{nicefrac}       %
\usepackage{lipsum}		%
\usepackage{graphicx}
\usepackage{natbib}
\usepackage{doi}

\usepackage[skip=0pt]{caption}
\usepackage{amsmath}
\usepackage{tikz}
\usepackage{float}
\usepackage{subcaption}
\usepackage{xcolor}         %
\usepackage{makecell}

\usepackage[acronym]{glossaries}

\newacronym{pls}{PLS}{Partial Least Squares}
\newacronym{pca}{PCA}{Principal Component Analysis}
\newacronym{pcr}{PCR}{Principal Component Regression}
\newacronym{ols}{OLS}{Ordinary Least Squares}
\newacronym{rr}{RR}{Ridge Regression}
\newacronym{en}{EN}{Elastic Net}
\newacronym{snr}{SNR}{Signal-to-Noise Ratio}
\newacronym{svd}{SVD}{Singular Value Decomposition}
\newacronym{lasso}{lasso}{Least Absolute Shrinkage and Selection Operator}
\newacronym{lfp}{LFP}{Lithium Iron Phosphate}
\newacronym{rss}{RSS}{Residual Sum of Squares}
\newacronym{rmse}{RMSE}{Root-Mean-Square Error}
\newacronym{nrmse}{NRMSE}{Normalized-Root-Mean-Square Error}
\newacronym{mape}{MAPE}{Mean-Absolute-Percentage Error}
\newacronym{nn}{NN}{Neural Network}

\usepackage{setspace} 
\title{Short Communication: \\ Interpretation of High-Dimensional Regression Coefficients by Comparison with Linearized Compressing Features}%

\date{} 					%

\author{ 
    Joachim Schaeffer\\
    Control and Cyber-Physical Systems Laboratory \\
    Technical University of Darmstadt, Germany \\
    \And
    \hspace{1em}Jinwook Rhyu\\ 
    \hspace{1em}Massachusetts Institute of Technology\hspace{1em}\\
    Cambridge, MA, USA\\
    \And
    Robin Droop\\ 
    \hspace{1.5em} Technical University of Munich, Germany \hspace{1.5em} \\ 
    \And
    Rolf Findeisen\\
    Control and Cyber-Physical Systems Laboratory \\
    Technical University of Darmstadt, Germany \\
    \And
    Richard D. Braatz\\
    Massachusetts Institute of Technology\\
    Cambridge, MA, USA\\
    \texttt{braatz@mit.edu}
}

\renewcommand{\headeright}{}
\renewcommand{\undertitle}{}
\renewcommand{\shorttitle}{Short Communication: Interpretation of High-Dimensional Regression by Comparison with Linearized Features}

\hypersetup{
pdftitle={Short Communication: Interpretation of High-Dimensional Regression by Comparison with Linearized Features},
pdfauthor={Joachim Schaeffer},
}

\begin{document}

\maketitle

\onehalfspacing
\begin{abstract}
  Linear regression is often deemed inherently interpretable; however, challenges arise for high-dimensional data. We focus on further understanding how linear regression approximates nonlinear responses from high-dimensional functional data, motivated by predicting cycle life for lithium-ion batteries.
  We develop a linearization method to derive feature coefficients, which we compare with the closest regression coefficients of the path of regression solutions.
  We showcase the methods on battery data case studies where a single nonlinear compressing feature, \(g\colon \mathbb{R}^p \to \mathbb{R}\), is used to construct a synthetic response, \(\mathbf{y} \in \mathbb{R}\). 
  This unifying view of linear regression and compressing features for high-dimensional functional data helps to understand (1) how regression coefficients are shaped in the highly regularized domain and how they relate to linearized feature coefficients and (2) how the shape of regression coefficients changes as a function of regularization to approximate nonlinear responses by exploiting local structures.
\end{abstract}

\newpage
\section{Introduction}
Interpretable machine learning methods aim to turn data into insights and knowledge for decision-making \cite{verdonck2021feature_eng, feature_eng_alma, domingos2012few}. Linear regression provides simple models that allow interpretation of how inputs \(\mathbf{X} \in \mathbb{R}^{n \times p}\) affect responses \(\mathbf{y} \in \mathbb{R}^n\) when the input dimension \(p\) is smaller than the sample size \(n\) \cite{hastie2009elements, gareth2021introduction}. However, for many problems in domains such as chemical engineering, chemistry, and biology, the input dimension is much larger than the sample size, \(p \gg n\) \cite{lavadeschafferbraatz2022}. Then, multicollinearity is inherently present because any column can be reconstructed by a linear combination of other columns, giving rise to the nullspace \cite{LA_GStrang, nullspaceschafferbraatz2024}. Regression methods commonly used in high dimensions, such as \gls{rr} and \gls{pls}, learn coefficients that are orthogonal to the nullspace by construction \cite{nullspaceschafferbraatz2024}, affecting interpretability. For data with a functional structure, the fused lasso can improve interpretability \cite{nullspaceschafferbraatz2024} but might affect model accuracy. Another research area is concerned with functional data analytics \cite{FunctionalDataAnalysis} and improving interpretability there \cite{interpretability_function_data_anlaysis}.

We are motivated by the prediction of lithium-ion battery cycle life \cite{schaeffer2024cycle, severson2019data} based on data collected from early-stage cycling. Voltage and current were continuously measured with high resolution \((p=1000)\), but the total number of batteries tested was small \((n=124\)), leading to high-dimensional data with functional structure. The variance, a single nonlinear compressing feature, displayed high predictive power for cycle life \cite{severson2019data}.\footnote{The term \textit{compressing feature} refers to transformations of data to produce a scalar, \(g\colon \mathbb{R}^p \to \mathbb{R}\).} A different approach is to use linear regression directly on high-dimensional data, leading to comparable prediction performance \cite{nullspaceschafferbraatz2024, schaeffer2024cycle}.

In past work, we analyzed how the nullspace shapes regression coefficients and how the nullspace can be used to improve interpretability \cite{nullspaceschafferbraatz2024}. Here, we are interested in further understanding how regression coefficients, \(\boldsymbol{\beta} \in \mathbb{R}^{p}\), learn a mapping to a response, \(\mathbf{y}\), that is a nonlinear response of the input data, \(\mathbf{X}\). 
To investigate this question, we construct synthetic nonlinear responses \(\mathbf{y}\) from lithium-ion battery data \cite{severson2019data} using the compressing feature $g$. We then derive linearized feature coefficients, \( \boldsymbol{\beta}_{\text{T1}}\), by applying the first-order Taylor expansion to $g$. Subsequently, we analyze the path of regression solutions and compare the regression coefficients closest to the feature coefficients with the feature coefficients. This perspective helps to improve the understanding of the highly regularized regression domain in high dimensions.
\section{Feature Linearization}
Linear regression models can be written as \((\mathbf{y}-\overline{y}\mathbf{1}) = (\mathbf{X}-\mathbf{1}\overline{\mathbf{x}}^\top)\boldsymbol\beta + \epsilon\) for a data set, and as 
\begin{align}
    (y_i-\overline{y}) &= (\mathbf{x}_i-\overline{\mathbf{x}})^\top\boldsymbol\beta + \epsilon_i,
    \label{eq:lin_reg_vec}
\end{align}
for a single sample $i$, where 
\(\mathbf{y} \in \mathbb{R}^n\) is a vector of observed responses, \(\overline{y} \in \mathbb{R} \) is the mean of the responses, \(\boldsymbol \epsilon \in \mathbb{R}^n\) is the model error, \(\mathbf{1} \in \mathbb{R}^n\) is a vector of ones, \(\mathbf{x}_i \in \mathbb{R}^{p}\) contains the data of sample \(i\), and \(\overline{\mathbf{x}}\in\mathbb{R}^p\) contains the column means of the data \(\mathbf{X} \in \mathbb{R}^{n \times p}\). Furthermore, \(\boldsymbol\beta \in \mathbb{R}^p\) are the regression coefficients, implying that they are estimated by regression.
Here, we are concerned with high-dimensional data with \(n\) samples and \(p\) measurements made over a continuous domain, and therefore \(p \gg n\). Such discrete measurements of an assumed smooth underlying process are also called {\em functional data} \cite{FunctionalDataAnalysis}. 
Additionally, \eqref{eq:lin_reg_vec} shows that regression coefficients are gradients with respect to the input \cite{hastie2009elements}.

We want to compare nonlinear compressing features, \(g\colon \mathbb{R}^p \to \mathbb{R}\), with regression coefficients. In the following, we linearize \(g\) and construct \textit{feature coefficients}.
We make three assumptions: Assumption 1:~There exists a nonlinear compressing feature, \(g\), that maps the data and the response, i.e., \(g(\mathbf{x}_i) = y_i + \epsilon_i^*\), where \(\epsilon_i^*\) is the error of the ground truth model (irreducible error). Assumption 2:~For simplicity, we set \(\epsilon_i^*=0 \ \forall i \in \{1, \dots{}, n\}\), but relax this assumption later. Assumption 3: \(g\) is differentiable. 

The first-order Taylor series approximation for functions with multiple inputs \cite{BirkhaeuserDistributions, pdoHoernmander}, applied to \(g\) and evaluated at \( \overline{\mathbf{x}}\), is
\begin{align}
    g_{\text{T1}}(\mathbf{x}_i;\overline{\mathbf{x}}) &= g(\overline{\mathbf{x}}) + (\mathbf{x}_i - \overline{\mathbf{x}})^\top\nabla g(\overline{\mathbf{x}}) = z_i. 
    \label{eq:combined_tay_lin_pre}
\end{align}
where \(\nabla g(\mathbf{\overline{\mathbf{x}}}):= \partial g/ \partial \mathbf{x}|_{\mathbf{x}=\overline{\mathbf{x}}}\) is the gradient of \(g\) evaluated at \(\overline{\mathbf{x}}\),  and \(z_i\) is the linearized feature corresponding to sample $i$. Rearranging \eqref{eq:combined_tay_lin_pre} yields
\begin{align}
    z_i - g(\overline{\mathbf{x}}) &= (\mathbf{x}_i - \overline{\mathbf{x}})^\top\nabla g(\overline{\mathbf{x}}).
    \label{eq:t1_rearragned}
\end{align}
The linearized feature \(z_i\) is an estimate for \(y_i\). Inserting \(z_i + \epsilon_{i, \text{T1}}^{g} = y_i\), where \(\epsilon_{i, \text{T1}}^{g}\) is the linearization error of \(g\) corresponding to sample $i$, into \eqref{eq:t1_rearragned} yields
\begin{align}
    y_i - g(\overline{\mathbf{x}}) &= (\mathbf{x}_i - \overline{\mathbf{x}})^\top\nabla g(\overline{\mathbf{x}}) + \epsilon_{i, \text{T1}}^g. \label{eq:t1_lin_reg}
\end{align}
Using Assumption 1 and the superscript \(g\) to indicate that the coefficients correspond to response that is subtracted by \(g(\overline{\mathbf{x}})\mathbf{1}\), we can reformulate the regression problem \eqref{eq:lin_reg_vec} to
\begin{align}
    (\mathbf{y}-g(\overline{\mathbf{x}})\mathbf{1}) &= (\mathbf{X}-\mathbf{1}\overline{\mathbf{x}}^\top)\boldsymbol\beta^g + \epsilon^g. \label{eq:lin_reg_g}
\end{align}
Comparing \eqref{eq:lin_reg_g} and \eqref{eq:t1_lin_reg} shows that both equations have the same structure.  %
However, the regression coefficients obtained by \eqref{eq:lin_reg_g} differ from \eqref{eq:lin_reg_vec} because
\begin{equation}
    \quad g(\overline{\mathbf{x}}) =: \bar{z} \approx \bar{y} 
    = \frac{1}{n}\sum_{i=1}^{n}g(\mathbf{x}_i)
    \label{eq:lin_assumption}
\end{equation}
might not be a good approximation. 
To correct for this, we suggest to estimate a scalar factor \(m\) via an additional regression step
\begin{align}
    (\mathbf{y}-\overline{y}\mathbf{1}) &= m(\mathbf{z}-\overline{z}\mathbf{1}) + \boldsymbol{\epsilon}_\text{T1},
    \label{eq:pred_n}
\end{align}
which also allows to relax Assumption 2. Here, \(\mathbf{z}\) is the vector of all \(z_i\) and \(\boldsymbol{\epsilon}_\text{T1}\) is the vector of linearized feature model errors. The scalar \(m\) is obtained by \gls{ols} and ensures that the feature coefficients estimate the true response. Combining \eqref{eq:t1_rearragned}, \eqref{eq:lin_assumption}, and \eqref{eq:pred_n} gives
\begin{equation}
    (\mathbf{y}-\overline{y}\mathbf{1}) = m(\mathbf{X}-\mathbf{1}\overline{\mathbf{x}}^\top)\nabla g(\overline{\mathbf{x}}) + \boldsymbol{\epsilon}_\text{T1}.
    \label{eq:derivative_inserted}
\end{equation}
We define the vector \(\boldsymbol{\beta}_{\text{T1}} := m\nabla g(\overline{\mathbf{x}})\) and refer to it as \textit{feature coefficients}. The feature coefficients can be compared with the path of regression solutions, as described in the next section.

\textit{Remark:} For simplicity, we work with a fixed training set in the following case studies and do not model the regression coefficients probabilistically. 

\paragraph{Regression Solution Path:}
Various linear models are suitable for regressing high-dimensional data.  The \gls{pls} model is commonly used in the chemometrics community, while \gls{rr}, lasso, and \gls{en} are more popular in the machine learning community. We do not consider the lasso and \gls{en} because their sparse regression coefficients inevitably differ from the structure of smooth derivatives that yield \(\boldsymbol\beta_\text{T1}\). Analyzing the lasso or \gls{en} would require further investigation and relaxation of assumptions. The \gls{pls} model has a single discrete regularization parameter, the number of components. \gls{rr} has a single continuous regularization parameter, \(\lambda\), making the path of regression solutions continuous. We want to explore the regression solution path and compare the regression coefficients \textit{closest} to the feature coefficients with the feature coefficients. 

While various norms can be used for calculating the distance between the regression and feature coefficients, we suggest estimating the regularization parameters based on the \(\ell^2\)-norm, 
\begin{equation}
\min_{\lambda} ||\boldsymbol\beta(\lambda) - \boldsymbol\beta_\text{T1}||_2^2,
\label{eq:reg_coef_hyp}
\end{equation}
due to its convenient properties. Inserting the closed-form solution of \gls{rr} in \eqref{eq:reg_coef_hyp} yields
\begin{equation}
    \min_{\lambda} ||(\mathbf{X^{\top}X + \lambda \mathbf{I}})^{-1}\mathbf{X^{\top}y} - \boldsymbol\beta_\text{T1}||_2^2,
\label{eq:min_reg_dist}
\end{equation}
which can be solved numerically. The regression coefficients of \gls{pls} and \gls{rr} are orthogonal to the nullspace of the data \(\mathbf{X}\) by construction \cite{nullspaceschafferbraatz2024}, 
However, due to the nullspace, there can be situations where \(||\boldsymbol\beta - \boldsymbol\beta_\text{T1}||_2^2\) is \textit{large} but the corresponding predictions are \textit{similar} (see Sec.\,\ref{sec:a_nullspace} and \cite{schaeffer2024cycle} for further discussion). 

\section{Case Studies}

\begin{figure}[htb!]
    \begin{subfigure}[b]{0.4735\linewidth}
    \centering
    \begin{tikzpicture}
    \node[anchor=south west, inner sep=0pt] at (0,0) (image1) {\includegraphics[width=\linewidth]{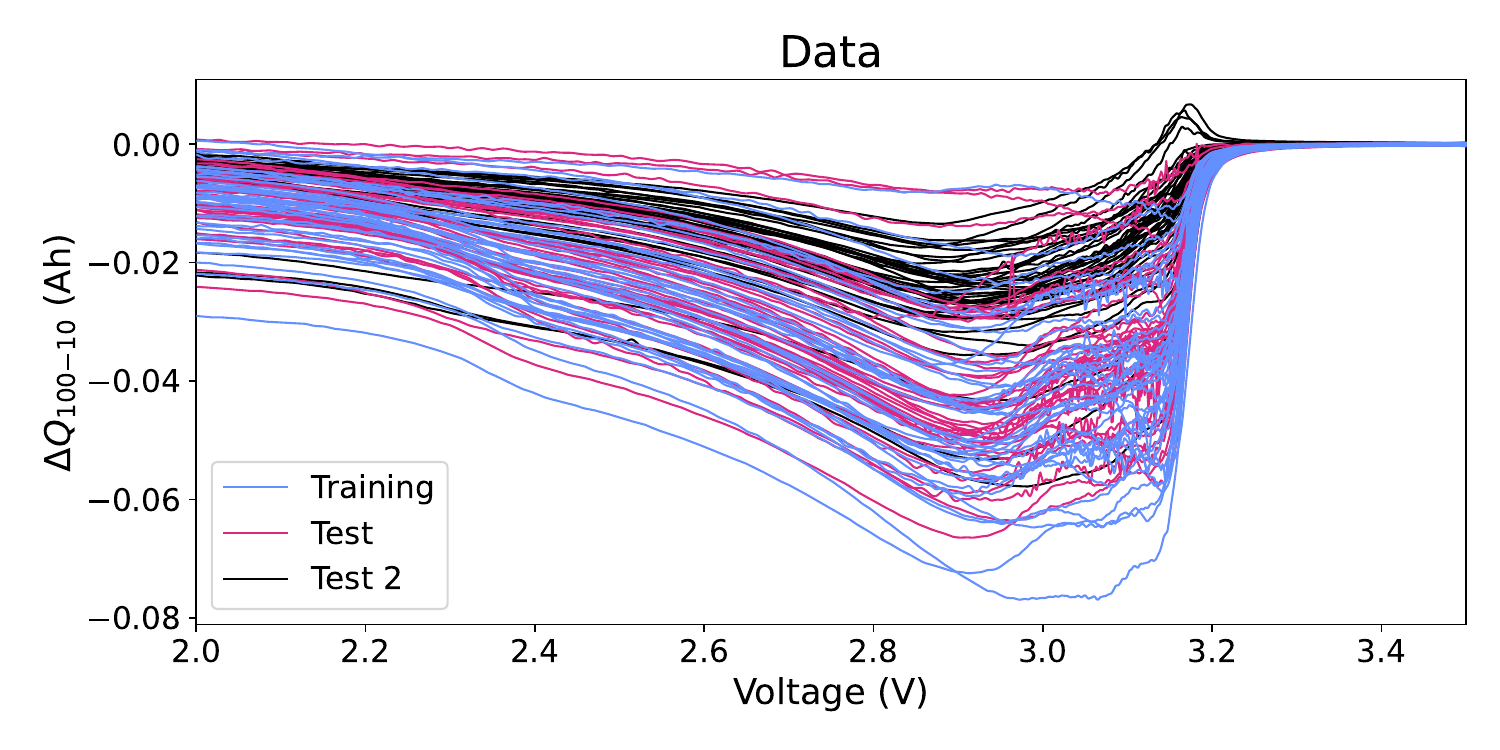}};
        \begin{scope}[x={(image1.south east)},y={(image1.north west)}]
        \node[fill=none] at (0.04,0.93) {\textbf{a)}};
        \end{scope}
    \end{tikzpicture} 
  \end{subfigure}%
  \begin{subfigure}[b]{0.4735\linewidth}
    \centering
    \begin{tikzpicture}
    \node[anchor=south west, inner sep=0pt] at (0,0) (image1) {\includegraphics[width=\linewidth]{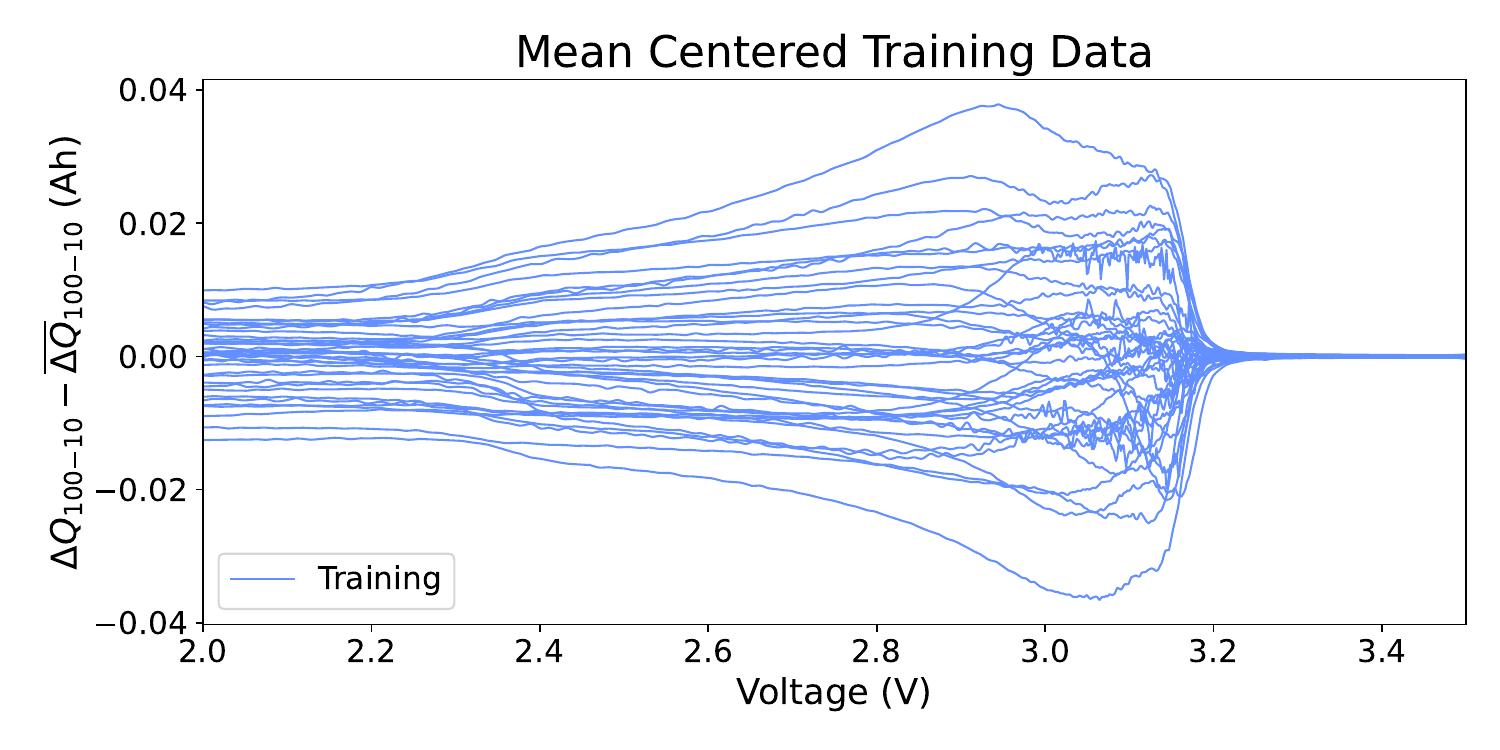}};
        \begin{scope}[x={(image1.south east)},y={(image1.north west)}]
        \node[fill=none] at (0.04,0.93) {\textbf{b)}};
        \end{scope}
    \end{tikzpicture} 
  \end{subfigure} 
\caption{Lithium-ion data from discharge cycles \cite{severson2019data}. a) Training, primary test, and secondary test data are plotted as curves. b) Mean-centered training data curves with one outlier removed.}
\label{fig:LFP_Data}
\end{figure}

The feature linearization method in this short note is motivated by cycle life prediction for lithium-ion batteries from early-stage cycling experiments where it was observed that a single compressing feature, the variance, had a good prediction performance, similar to the performance of regression in high dimensions (compare models in \cite{schaeffer2024cycle} and \cite{severson2019data}). Therefore, we use the lithium-ion battery data set \cite{severson2019data} for the case studies in this work. The data \(\mathbf{X} \in \mathbb{R}^{124 \times 1000}\) is the difference in discharge capacity of cycles 100 and 10 as a function of voltage (Fig.\,\ref{fig:LFP_Data}). The data set was partitioned into a training (\(\mathbf{X}^\text{train} \in \mathbb{R}^{40 \times 1000}\) after removing one outlier), primary test (\(\mathbf{X}^\text{test1} \in \mathbb{R}^{43 \times 1000}\)), and secondary test (\(\mathbf{X}^\text{test2}  \in \mathbb{R}^{40 \times 1000}\)) sets according to \cite{severson2019data}. To test our methodology, we only use the training set in the following. We did not z-score the data because the variance feature was applied to the non-z-scored data in \cite{severson2019data}. Furthermore, the noise in the data is heteroscedastic, and the signal-to-noise ratio drops significantly in the region 3.2--3.5\,V for physical reasons (see \cite{nullspaceschafferbraatz2024} for more details on z-scoring as a design choice). %
To study feature linearization, we construct synthetic responses \(\mathbf{y}\) by using two different compressing features. 

\paragraph{Case Study I:} The first case study uses the sum of squares \(g(\mathbf{x}_i) = \sum_{j=1}^p x_{i, j}^2 = y_i \) (Fig.\,\ref{fig:case_study1}). The feature coefficients of the sum-of-squares feature are the scaled mean of the data (compare Fig.\,\ref{fig:case_study1}a and Fig.\,\ref{fig:LFP_Data_all}d; the derivation is shown in Sec.\,\ref{sec:derivation_sos_der}). %
For large regularization, the regression coefficients of \gls{rr} approach zero over the entire domain (Fig.\,\ref{fig:case_study1}a) while remaining a shape that is similar to the standard deviation of the data (Fig.\,\ref{fig:LFP_Data_all}d). The \gls{rr} and \gls{pls} regression coefficients with minimal \(\ell^2\)-distance to the feature coefficients have a very similar shape as the feature coefficients. However, they do not match exactly due to the way how regularization shapes the regression coefficients. Lowering the regularization further by choosing the regularization value by 10-fold cross-validation according to the one-standard-error rule \cite{hastie2009elements, repeated_double_cv} increases the magnitude of regression coefficients.
Further decreasing the regularization by not using the one-standard-error rule leads to an even larger magnitude of the regression coefficients that change faster across the voltage input domain (Fig.\,\ref{fig:case_study1}).
\begin{figure}[H]
    \centering
    \begin{subfigure}[b]{0.4735\linewidth}
        \raggedleft
        \begin{tikzpicture}
            \node[anchor=south west, inner sep=0pt] at (0,0) (image1) {\includegraphics[trim={0cm, 0.2cm, 0cm, 1cm}, clip, width=\linewidth]{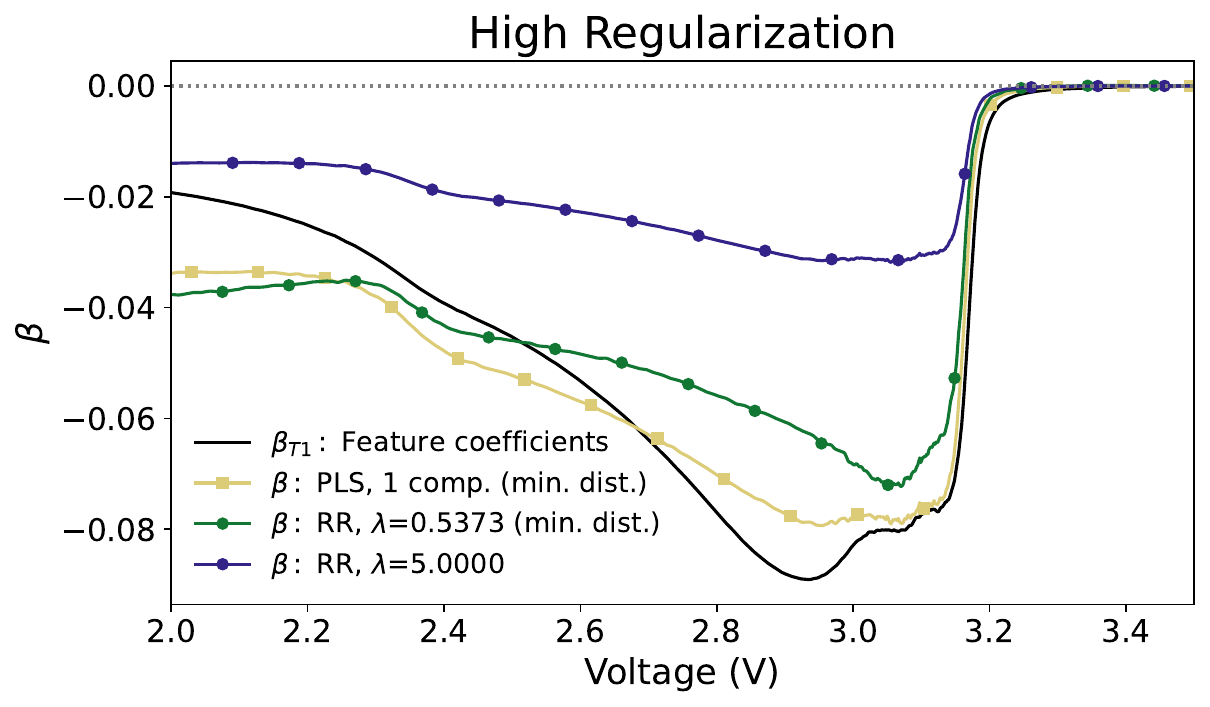}};
            \begin{scope}[x={(image1.south east)},y={(image1.north west)}]
            \node[fill=none, anchor=south west] at (0.05,1) {\footnotesize	 \textbf{a)} High regularization};
            \end{scope}
        \end{tikzpicture} 
    \end{subfigure}
    \begin{subfigure}[b]{0.4735\linewidth}
        \raggedleft
        \begin{tikzpicture}
            \node[anchor=south west, inner sep=0pt] at (0,0) (image1) {\includegraphics[trim={0cm, 0.2cm, 0cm, 1cm}, clip, width=0.98\linewidth]{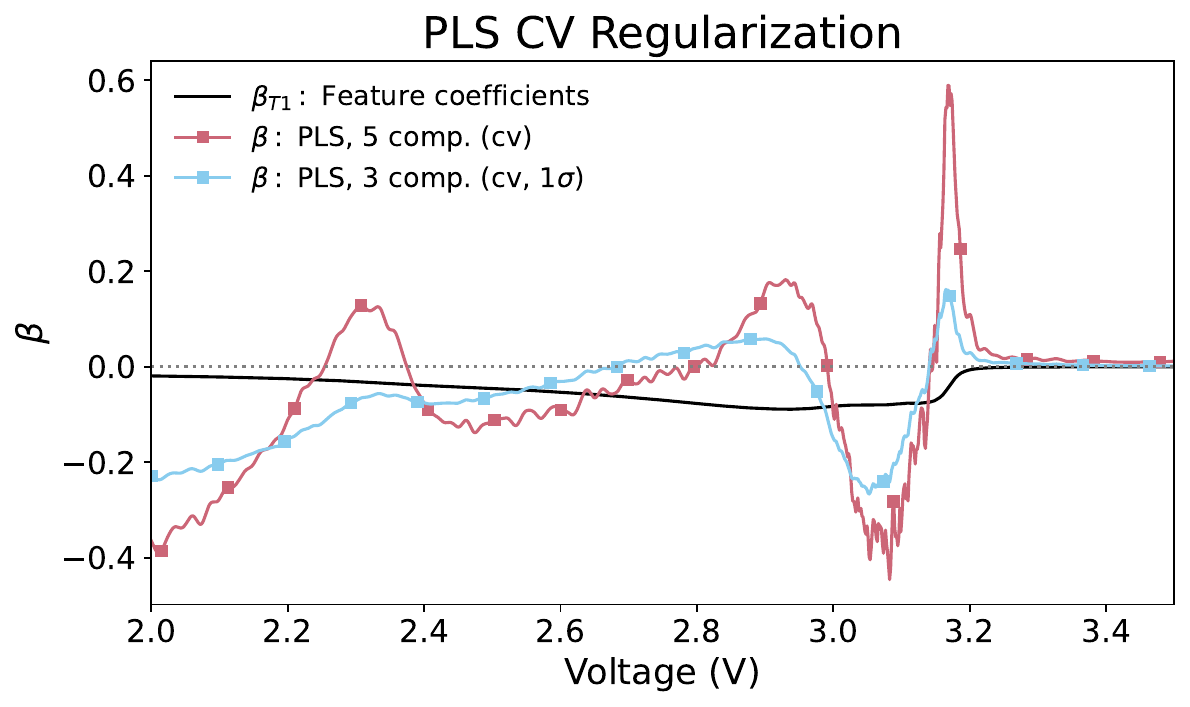}};
            \begin{scope}[x={(image1.south east)},y={(image1.north west)}]
            \node[fill=none, anchor=south west] at (0.05,1) {\footnotesize	 \textbf{b)} PLS CV regularization};
            \end{scope}
        \end{tikzpicture} 
    \end{subfigure}%
    \begin{subfigure}[b]{0.4735\linewidth}
        \raggedleft
        \begin{tikzpicture}
            \node[anchor=south west, inner sep=0pt] at (0,0) (image1) {\includegraphics[trim={0cm, 0cm, 0cm, 1cm}, clip, width=0.96\linewidth]{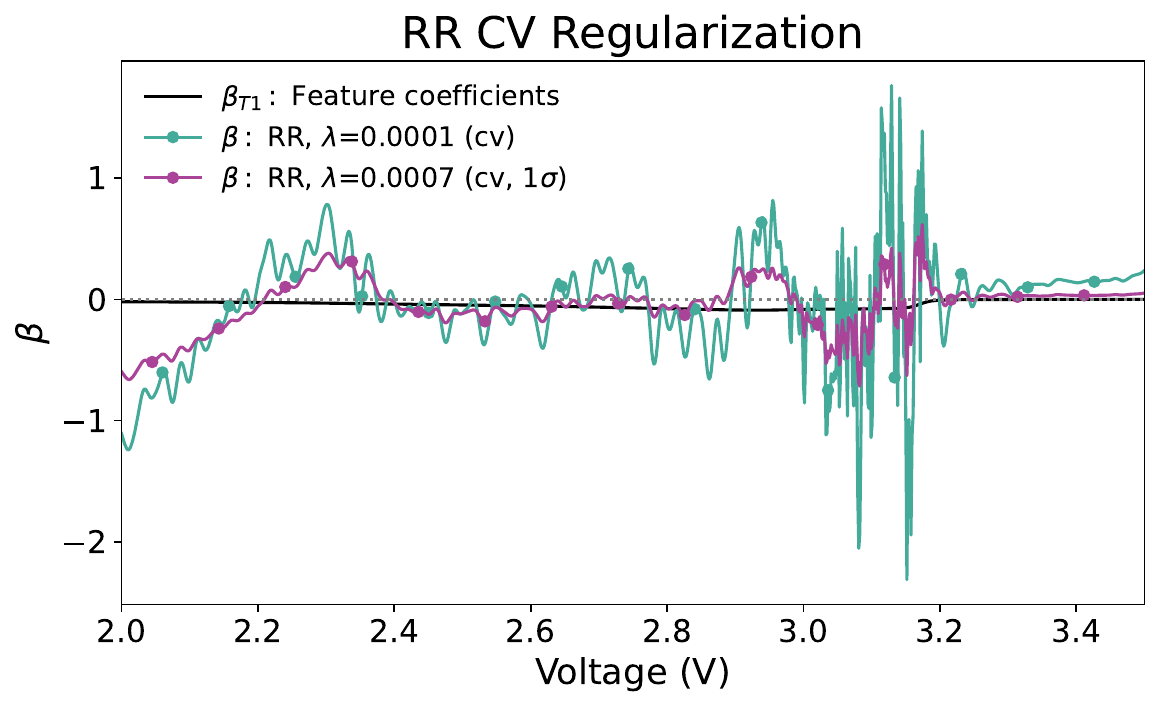}};
            \begin{scope}[x={(image1.south east)},y={(image1.north west)}]
            \node[fill=none, anchor=south west] at (0.05,1) {\footnotesize	 \textbf{c)} RR CV regularization};
        \end{scope}
    \end{tikzpicture} 
    \end{subfigure}
    \vspace{0.2cm}
    \caption{First case study with the sum-of-squares response. a) 
    \gls{pls} and \gls{rr} with high regularization and sum-of-squares feature coefficients. b) \gls{pls} regression coefficients obtained by cross-validation and sum-of-squares feature coefficients. c) \gls{rr} regression coefficients obtained by cross-validation and sum-of-squares feature coefficients.}
    \label{fig:case_study1}
\end{figure}
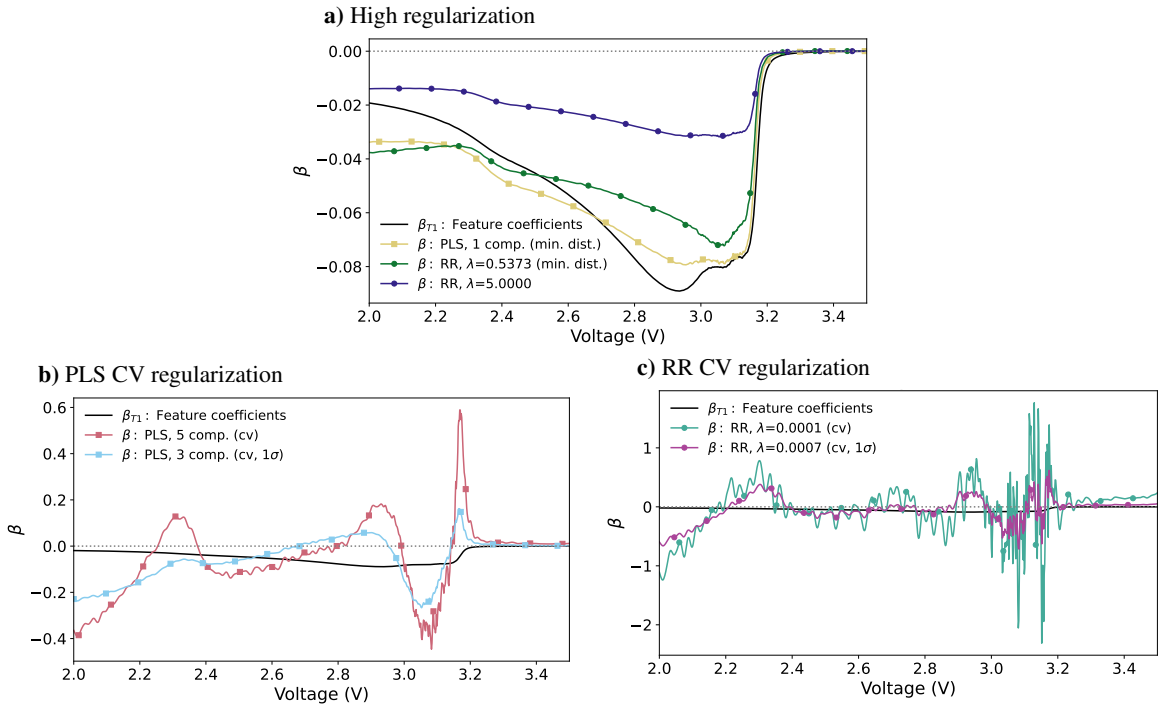
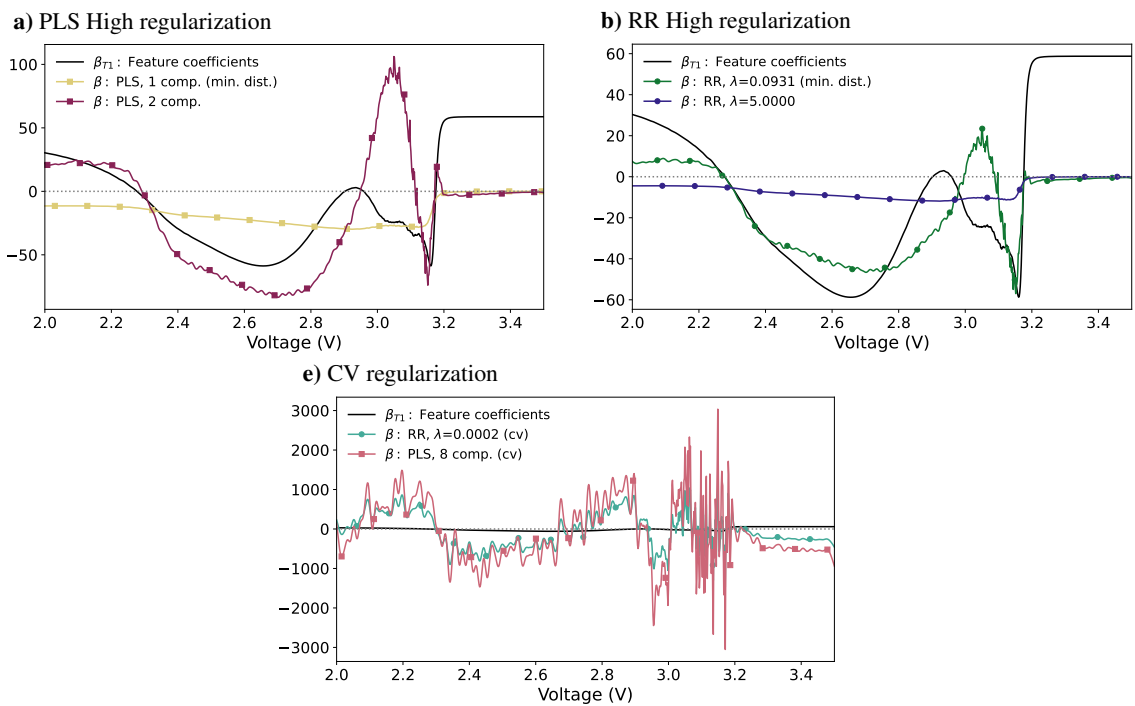
\begin{figure}[H]
    \centering
    \begin{subfigure}[b]{0.4735\linewidth}
        \raggedleft
        \begin{tikzpicture}
            \node[anchor=south west, inner sep=0pt] at (32,0) (image1) {\includegraphics[trim={1cm, 0.2cm, 0cm, 1cm}, clip, width=0.929\linewidth]{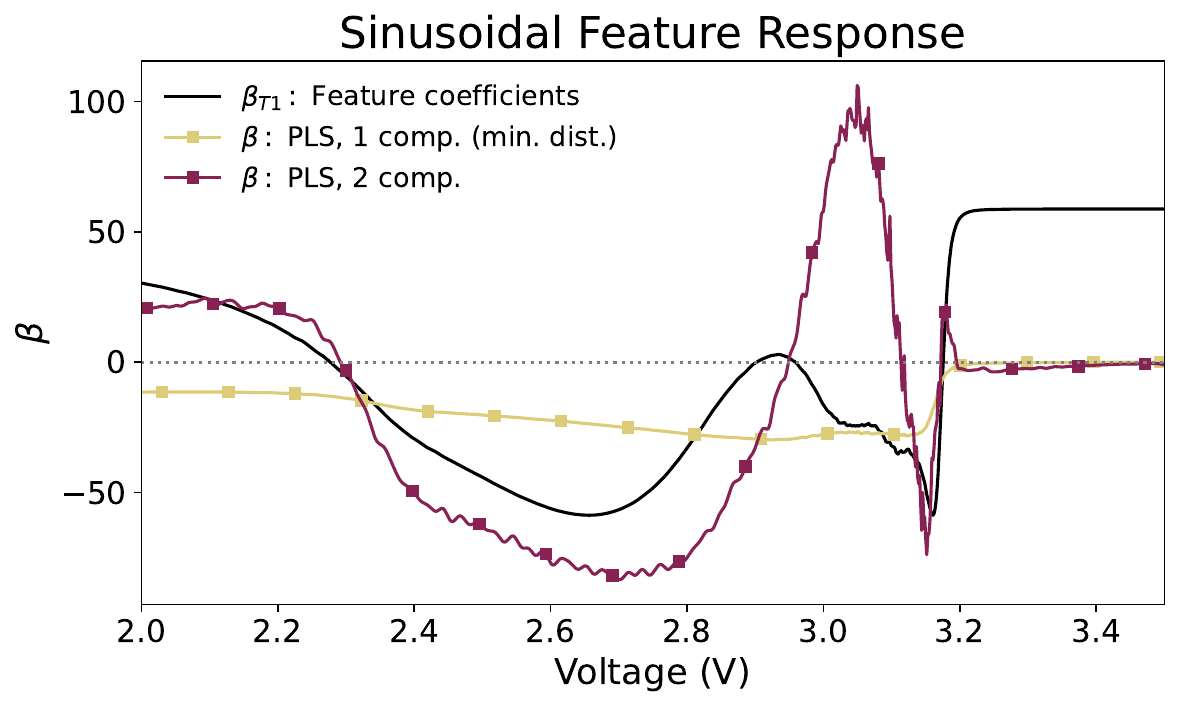}};
            \begin{scope}[x={(image1.south east)},y={(image1.north west)}]
            \node[fill=none, anchor=south west] at (0,1) {\footnotesize	 \textbf{a)} PLS High regularization};
            \end{scope}
        \end{tikzpicture} 
    \end{subfigure}%
    \begin{subfigure}[b]{0.4735\linewidth}
        \raggedleft
            \begin{tikzpicture}
            \node[anchor=south west, inner sep=0pt] at (32,0) (image1) {\includegraphics[trim={1cm, 0.2cm, 0cm, 1cm}, clip, width=0.93\linewidth]{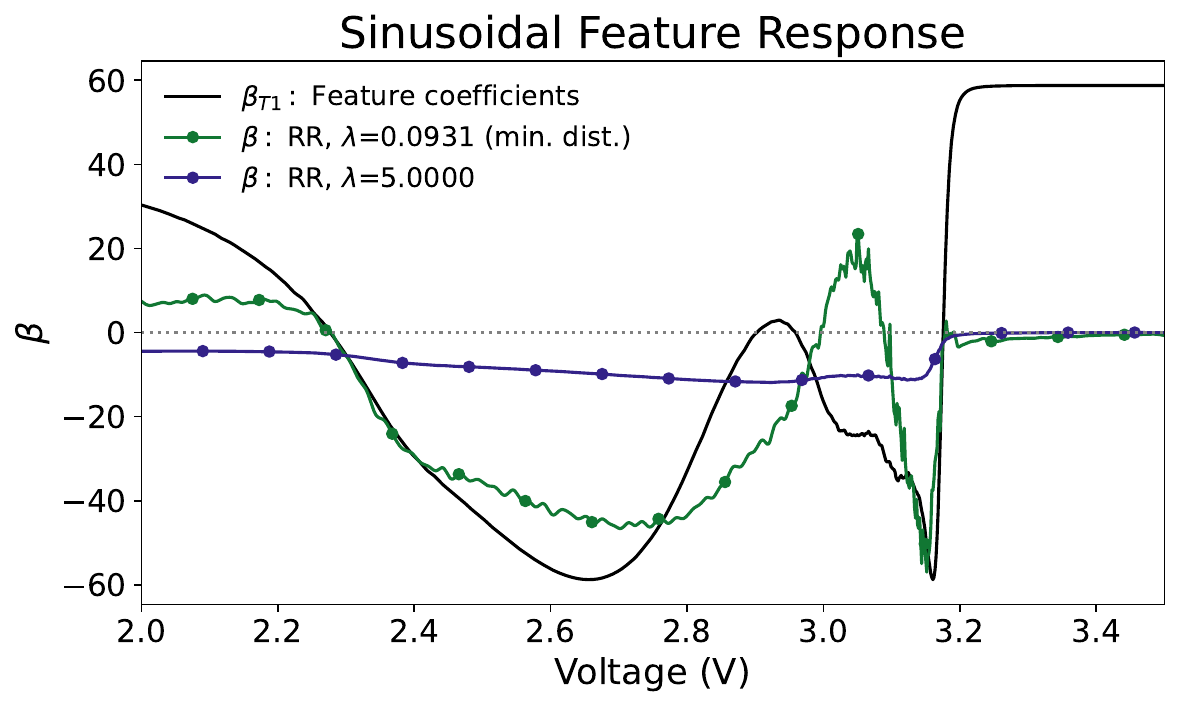}};
            \begin{scope}[x={(image1.south east)},y={(image1.north west)}]
            \node[fill=none, anchor=south west] at (0,1) {\footnotesize	 \textbf{b)} RR High regularization};
            \end{scope}
        \end{tikzpicture} 
    \end{subfigure}
    
    \begin{subfigure}[b]{0.4735\linewidth}
        \raggedleft
        \begin{tikzpicture}
            \node[anchor=south west, inner sep=0pt] at (0,0) (image1) {\includegraphics[trim={1cm, 0cm, 0cm, 1cm}, clip, width=0.96\linewidth]{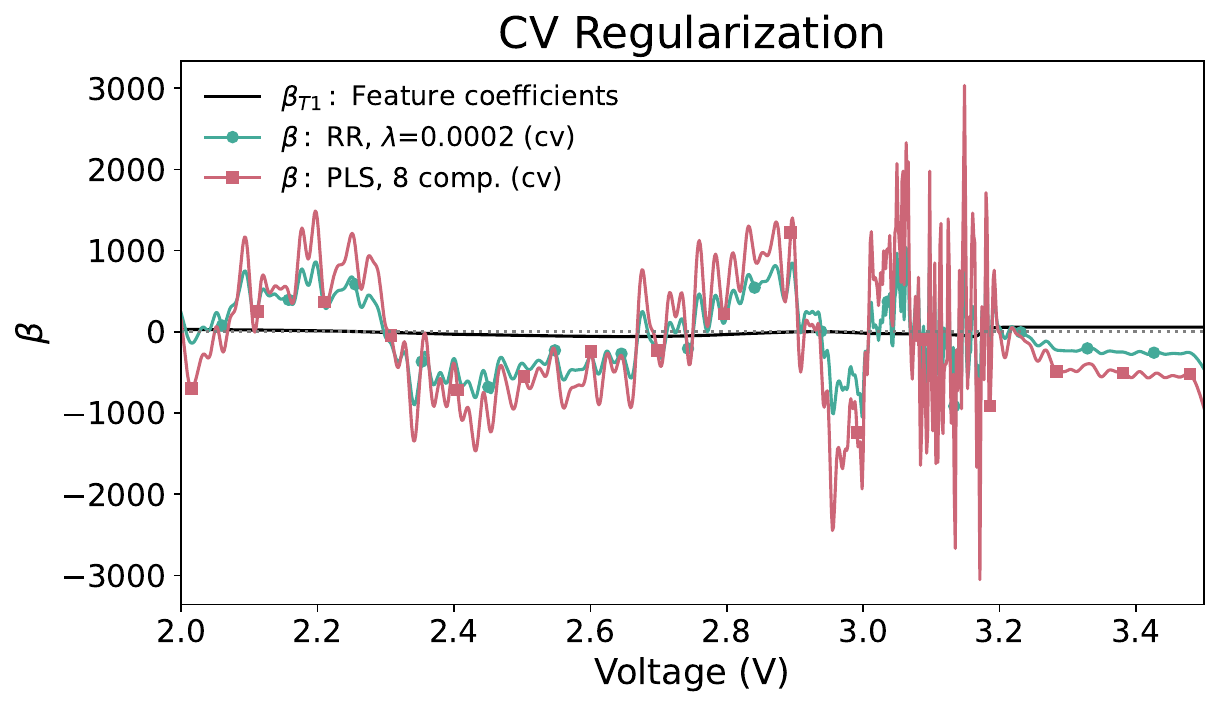}};
            \begin{scope}[x={(image1.south east)},y={(image1.north west)}]
            \node[fill=none, anchor=south west] at (0.03,1) {\footnotesize	 \textbf{e)} CV regularization};
            \end{scope}
        \end{tikzpicture} 
    \end{subfigure} 
    \vspace{0.2cm}
    \caption{Second case study with the sinusoidal response. a) 
    \gls{pls} high regularization and sinusoidal feature coefficients. b) \gls{rr} high regularization and sinusoidal feature coefficients. c) \gls{rr} and \gls{pls} regression coefficients obtained by cross-validation and sinusoidal feature coefficients.}
    \label{fig:case_study2}
\end{figure}
\paragraph{Case Study II:}
The second case study considers a sinusoidal feature, \(g(\mathbf{x}_i) =  \sum_{j=1}^p \sin\!\left(\frac{2\pi}{0.06}x_{i, j}\right)=y_i\). The sinusoidal feature coefficients have a very different shape than the variance of the data (Fig.\,\ref{fig:case_study2}a and Fig.\,\ref{fig:LFP_Data_all}). Similarly to the sum-of-squares feature, the shape of the highly regularized regression coefficients of \gls{rr} and \gls{pls} are dominated by the directions with high variance (Figs.\,\ref{fig:case_study2}ab). The \gls{pls} model with one component has minimum \(\ell^2\)-distance to the feature coefficients but is still dominated by the variance of the data and does not yet pick up the characteristic shape of the sinusoidal feature coefficients. The \gls{pls} model with two components shows a similar shape as the feature coefficients but overshoots them (Fig.\,\ref{fig:case_study2}a). For \gls{rr}, the minimum distance regression coefficients are closer
to the feature coefficients in terms of \(\ell^2\)-norm than the \gls{pls} coefficients and also have the characteristic shape of the feature coefficients (Fig.\,\ref{fig:case_study2}b). Choosing the regularization value by 10-fold cross-validation
yields regression coefficients with a larger magnitude, exploiting local structures (Fig.\,\ref{fig:case_study2}c), phenomenologically similar to the sum-of-squares case study. 

Conceptually, a linear model is expected first to capture the ``key'' directions of the data, subject to the regularization constraint. Afterward, the model tries to approximate the nonlinearity linearly in high dimensions. Although the linear model might not fully capture nonlinearity, the case studies in this work and \cite{lavadeschafferbraatz2022} suggest that linear models can perform well on high-dimensional data when a limited amount of nonlinearity is present. Furthermore, these case studies demonstrate how the regression coefficients exploit local structures to approximate nonlinearity.

\section{Conclusion}
The key contribution of this work is to further the understanding of how the shape of regression coefficients for high-dimensional functional data is affected by the degree of regularization. We developed a linearization method to compare regression coefficients with feature coefficients. We show how the variance of the data dominates regression coefficients obtained by very strong regularization. 
The regression solution path contains regression coefficients similar to the linearized true mapping of input data and response (i.e., feature coefficients) for strong regularization. Furthermore, reducing the regularization to the regularization parameter estimated by cross-validation yields regression coefficients that change faster in the input data domain, exploiting the data's local structures to approximate the nonlinear response linearly in high dimensions. \\
While we used synthetic responses for case studies, in reality, there is no ideal feature, and good features that might exist are difficult to discover. Systemically designing features (e.g., \cite{rhyu2024systematic}) that generalize well and are interpretable is an active area of research. We hope that, ultimately, the ideas in this work can support such efforts.

\section*{Acknowledgements and Funding}
We would like to thank Dr. Eric Lenz for very helpful feedback.

We acknowledge base funding from the Technical University of Darmstadt. Furthermore, this work was refined during Joachim Schaeffer's time at the Massachusetts Institute of Technology, for which we acknowledge financial support by a fellowship within the IFI program of the German Academic Exchange Service (DAAD), funded by the Federal Ministry of Education and Research (BMBF).

\section*{Data}
The \gls{lfp} data are available at \url{https://data.matr.io/1/} under the terms of CC BY 4. 

\clearpage
\appendix

\section{Appendix}
\setcounter{figure}{0} 
\counterwithin{figure}{section}
\subsection{Lithium-ion Battery Data Set}
\label{sec:si_lion_data}
The \gls{lfp} battery dataset used in this article contains cycling data for 124 batteries and was initially published with \cite{severson2019data}. Each battery is charged with a fixed charging protocol. Each charging protocol was applied to multiple cells thus the number of unique charging protocols is smaller than the number of batteries. The discharge was constant at 4C (corresponding to a 15-minute discharge) and identical for all cells \cite{severson2019data}. The batteries responded to the stress of the experiment by showing different capacity fade curves. 
The objective in  \cite{severson2019data} was to develop a model that learns signs of battery degradation from early discharge cycles and uses this information to predict the number of cycles until the battery capacity drops below \(80\%\) of its nominal capacity (i.e., the cycle life). Ref. \cite{severson2019data} found that significant information about cell degradation, and thus a source of characteristics, is the difference between discharge vectors that map the integral of current over time to the voltage of the cell for two distinct cycles. The full data matrix containing training, primary, and secondary test set \(\mathbf{X} \in  \mathbb{R}^{123 \times 1000}\) has a large number of columns due to the high measurement resolution of current over voltage. More technical details can be found in \cite{severson2019data}. Similarly shaped high-dimensional data often appear in chemical and biological systems due to measurements over a continuous domain.
\begin{figure}[h]
    \centering
    \includegraphics[width=0.97\linewidth]{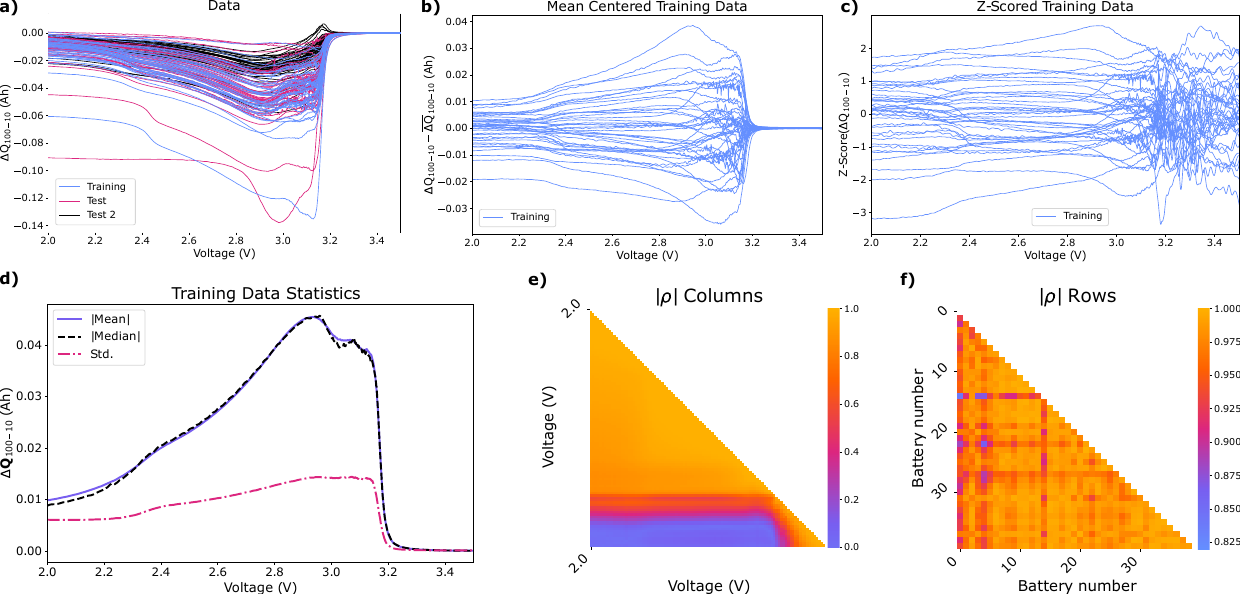}
    \vspace{4ex}
    \caption{Lithium-ion data from discharge cycles \cite{severson2019data}. a) Training, primary test, and secondary test data plotted as curves. b) Mean-centered training data curves with one outlier removed, c) z-scored training data curves with one outlier removed, d) training data statistics, e) Pearson correlation coefficient training data columns, f) Pearson correlation coefficients of training data rows.}
    \label{fig:LFP_Data_all}
\end{figure}
\newpage
\subsection{Derivation of Feature Coefficients for the Sum-Of-Squares Feature}
\label{sec:derivation_sos_der}
For the sum-of-squares case study, the nonlinear response \(\mathbf{y}\) is
\begin{align}
    y_i = \sum_{j=1}^p x_{i, j}^2.
    \label{eq:sos}
\end{align}
The feature coefficients are
\begin{align}
 \frac{\partial f(\mathbf{x}_i)}{\partial x_k}  
    &=  2x_{i,k} \notag\\
   \nabla f(\overline{\mathbf{x}}) &= 2 \mathbf{\overline{x}} \notag\\
   \boldsymbol{\beta}_{\text{T1}} &= m\nabla f(\overline{\mathbf{x}}) = 2m \mathbf{\overline{x}},
    \label{eq:sos_coef}
\end{align}
where \(\mathbf{\overline{x}} = \frac{1}{n} \sum_{i=1}^n \mathbf{x}_i\) is the vector containing column averages of the data matrix. The feature coefficients for the sum-of-squares response are, therefore, the scaled column mean.
Deriving the feature coefficients for a large number of features is tedious and error-prone, so the software accompanying the article implements the derivative via automatic differentiation.

\subsection{Nullspace Analysis, Sinusoidal Response, PLS Regression }
\label{sec:a_nullspace}
In high dimensions, the nullspace allows very different-looking regression coefficients to yield similar predictions \cite{nullspaceschafferbraatz2024}. Therefore, as an alternative to \eqref{eq:min_reg_dist}, 
\begin{equation}
\min_{\lambda} ||\mathbf{X}\boldsymbol\beta(\lambda) - \mathbf{X}\boldsymbol\beta_\text{T1}||_2^2
\label{eq:min_y_diff}
\end{equation}
could be used to yield regression coefficients with predictions closest to the feature coefficient predictions. However, applying \eqref{eq:min_y_diff} might make it more difficult to compare the resulting coefficients. %

\newpage
\bibliography{references}

\begin{thebibliography}{16}
\providecommand{\natexlab}[1]{#1}
\providecommand{\url}[1]{\texttt{#1}}
\expandafter\ifx\csname urlstyle\endcsname\relax
  \providecommand{\doi}[1]{doi: #1}\else
  \providecommand{\doi}{doi: \begingroup \urlstyle{rm}\Url}\fi

\bibitem[Verdonck et~al.(2024)Verdonck, Baesens, {\'O}skarsd{\'o}ttir, and
  vanden Broucke]{verdonck2021feature_eng}
Tim Verdonck, Bart Baesens, Mar{\'\i}a {\'O}skarsd{\'o}ttir, and Seppe vanden
  Broucke.
\newblock Special issue on feature engineering editorial.
\newblock \emph{Machine learning}, 113\penalty0 (7):\penalty0 3917--3928, 2024.

\bibitem[Zheng and Amanda(2018)]{feature_eng_alma}
Alice Zheng and Casari Amanda.
\newblock \emph{Feature Engineering for Machine Learning: Principles and
  Techniques for Data Scientists}.
\newblock O'Reilly, Beijing, 2018.

\bibitem[Domingos(2012)]{domingos2012few}
Pedro Domingos.
\newblock A few useful things to know about machine learning.
\newblock \emph{Communications of the ACM}, 55\penalty0 (10):\penalty0 78--87,
  2012.

\bibitem[Hastie et~al.(2009)Hastie, Tibshirani, Friedman, and
  Friedman]{hastie2009elements}
Trevor Hastie, Robert Tibshirani, Jerome~H. Friedman, and Jerome~H. Friedman.
\newblock \emph{The Elements of Statistical Learning: Data Mining, Inference,
  and Prediction}.
\newblock Springer, New York, 2009.

\bibitem[Gareth et~al.(2021)Gareth, Daniela, Trevor, and
  Robert]{gareth2021introduction}
James Gareth, Witten Daniela, Hastie Trevor, and Tibshirani Robert.
\newblock \emph{An Introduction to Statistical Learning: with Applications in
  R}.
\newblock Springer, New York, 2021.

\bibitem[Schaeffer and Braatz(2022)]{lavadeschafferbraatz2022}
Joachim Schaeffer and Richard~D. Braatz.
\newblock Latent variable method demonstrator -- {Software} for understanding
  multivariate data analytics algorithms.
\newblock \emph{Computers \& Chemical Engineering}, 167:\penalty0 108014, 2022.

\bibitem[Strang(2016)]{LA_GStrang}
Gilbert Strang.
\newblock \emph{Introduction to Linear Algebra}.
\newblock Cambridge Press, Wellesley, Massachusetts, fifth edition, 2016.

\bibitem[Schaeffer et~al.(2024{\natexlab{a}})Schaeffer, Lenz, Chueh, Bazant,
  Findeisen, and Braatz]{nullspaceschafferbraatz2024}
Joachim Schaeffer, Eric Lenz, William~C. Chueh, Martin~Z. Bazant, Rolf
  Findeisen, and Richard~D. Braatz.
\newblock Interpretation of high-dimensional linear regression: Effects of
  nullspace and regularization demonstrated on battery data.
\newblock \emph{Computers \& Chemical Engineering}, 180:\penalty0 108471,
  2024{\natexlab{a}}.

\bibitem[Ramsay and Silverman(2005)]{FunctionalDataAnalysis}
J.~O. Ramsay and B.~W. Silverman.
\newblock \emph{Functional Data Analysis}.
\newblock Springer, New York, second edition, 2005.

\bibitem[James et~al.(2009)James, Wang, and
  Zhu]{interpretability_function_data_anlaysis}
Gareth~M. James, Jing Wang, and Ji~Zhu.
\newblock Functional linear regression that’s interpretable.
\newblock \emph{The Annals of Statistics}, 37\penalty0 (5A):\penalty0
  2083--2108, 2009.

\bibitem[Schaeffer et~al.(2024{\natexlab{b}})Schaeffer, Galuppini, Rhyu,
  Asinger, Droop, Findeisen, and Braatz]{schaeffer2024cycle}
Joachim Schaeffer, Giacomo Galuppini, Jinwook Rhyu, Patrick~A. Asinger, Robin
  Droop, Rolf Findeisen, and Richard~D. Braatz.
\newblock Cycle life prediction for lithium-ion batteries: {Machine} learning
  and more.
\newblock In \emph{Proceedings of the American Control Conference}, pages
  763--768, 2024{\natexlab{b}}.

\bibitem[Severson et~al.(2019)Severson, Attia, Jin, Perkins, Jiang, Yang, Chen,
  Aykol, Herring, Fraggedakis, Bazant, Harris, Chueh, and
  Braatz]{severson2019data}
Kristen~A. Severson, Peter~M. Attia, Norman Jin, Nicholas Perkins, Benben
  Jiang, Zi~Yang, Michael~H. Chen, Muratahan Aykol, Patrick~K. Herring,
  Dimitrios Fraggedakis, Martin~Z. Bazant, Stephen~J. Harris, William~C. Chueh,
  and Richard~D. Braatz.
\newblock Data-driven prediction of battery cycle life before capacity
  degradation.
\newblock \emph{Nature Energy}, 4\penalty0 (5):\penalty0 383--391, 2019.

\bibitem[Duistermaat(2010)]{BirkhaeuserDistributions}
J.~J. Duistermaat.
\newblock \emph{{Distributions: Theory and Applications}}.
\newblock Birkhäuser, Boston, Massachusetts, 2010.

\bibitem[Hörmander(2003)]{pdoHoernmander}
Lars Hörmander.
\newblock \emph{{The Analysis of Linear Partial Differential Operators I:
  Distribution Theory and Fourier Analysis}}.
\newblock Springer, Berlin--Heidelberg, second edition, 2003.

\bibitem[Filzmoser et~al.(2009)Filzmoser, Liebmann, and
  Varmuza]{repeated_double_cv}
Peter Filzmoser, Bettina Liebmann, and Kurt Varmuza.
\newblock Repeated double cross validation.
\newblock \emph{Journal of Chemometrics}, 23\penalty0 (4):\penalty0 160--171,
  2009.

\bibitem[Rhyu et~al.(2024)Rhyu, Schaeffer, Li, Cui, Chueh, Bazant, and
  Braatz]{rhyu2024systematic}
Jinwook Rhyu, Joachim Schaeffer, Michael~L. Li, Xiao Cui, William~C. Chueh,
  Martin~Z. Bazant, and Richard~D. Braatz.
\newblock Systematic feature design for cycle life prediction of lithium-ion
  batteries during formation.
\newblock \emph{arXiv preprint arXiv:2410.07458}, 2024.

\end{thebibliography}

\end{document}